\documentclass[10pt,twocolumn,letterpaper]{article}

\usepackage{cvpr}
\usepackage{times}
\usepackage{epsfig}
\usepackage{graphicx}
\usepackage{algorithm}
\usepackage{algorithmic}
\usepackage{amsmath}
\usepackage{amssymb}
\usepackage{bm}

\newcommand{\INCIDENT}[1][1]{\hspace{#1\algorithmicindent}}

\usepackage[breaklinks=true,bookmarks=false]{hyperref}

\cvprfinalcopy 

\begin{document}

\title{AlignGAN: Learning to Align Cross-Domain Images with \\
Conditional Generative Adversarial Networks}

\author{Xudong Mao\\
Department of Computer Science\\
City University of Hong Kong\\
{\tt\small xudonmao@gmail.com}\\
\and
Qing Li\\
Department of Computer Science\\
City University of Hong Kong\\
{\tt\small itqli@cityu.edu.hk}
\and
Haoran Xie\\
Department of Mathematics and \\
Information Technology\\
The Education University of Hong Kong\\
{\tt\small hrxie2@gmail.com}
}

\maketitle

\begin{abstract}
Recently, several methods based on generative adversarial network (GAN) have been proposed for the task of aligning cross-domain images or learning a joint distribution of cross-domain images. One of the methods is to use conditional GAN for alignment. However, previous attempts of adopting conditional GAN do not perform as well as other methods. In this work we present an approach for improving the capability of the methods which are based on conditional GAN. We evaluate the proposed method on numerous tasks and the experimental results show that it is able to align the cross-domain images successfully in absence of paired samples. Furthermore, we also propose another model which conditions on multiple information such as domain information and label information. Conditioning on domain information and label information, we are able to conduct label propagation from the source domain to the target domain. A 2-step alternating training algorithm is proposed to learn this model.
\end{abstract}

\section{Introduction}
Generative Adversarial Networks (GAN) \cite{Goodfellow2014} has proven hugely successful for various computer vision tasks \cite{Isola2016,Ledig2016,Radford2015}. This paper addresses the problem of aligning cross-domain images or learning a joint distribution of cross-domain images \cite{Liu2016}. Early approaches \cite{Isola2016,Wang2012} for this problem require paired images from different domains, which limits the effectiveness of these approaches. Recently, CoGAN \cite{Liu2016} has been proposed, lifting the restriction of paired images. In particular, CoGAN couples two GANs in which two generators share the weights of the first several layers, which guides the two generators to generate aligned images.

In this paper, we introduce a model called AlignGAN for aligning cross-domain images, which is based on the conditional GAN \cite{Mirza2014}. Similar to CoGAN, our proposed AlignGAN is also able to align cross-domain images without paired images. The idea of using conditional GAN for alignment is to learn the domain-specific semantics by the conditioned domain vectors and to learn the shared semantics by the other latent vectors. However, as pointed out in literature \cite{Liu2016}, adopting conditional GAN directly will fail to align cross-domain images for some tasks. We find that determining which layers to be conditioned by domain vectors is critical to the performance. Our proposed AlignGAN is inspired by the following two ideas. First, for the generator, the highest level semantics of different domains should be similar. Thus we should not condition the domain vectors on the noise input layer of the generator. Second, for the discriminator, we should enhance the domain information signals to let the discriminator know which domain the images are from. The image input layer generates the strongest signal for the discriminator. Thus we should condition the domain vectors on the image input layer of the discriminator. We explore AlignGAN for many tasks including digits and negative digits, blond hair and black hair, and chairs and cars. Furthermore, AlignGAN is not limited to two domains and it can be extended to three or more domains by just adding more dimensions to the domain vectors as Figure \ref{fig:face}(a) shows.

Based on AlignGAN, we also propose another model that is conditioned on multiple information such as domain information and label information. Suppose we only have the label information of the source domain. By learning the label information from the source domain and aligning the images using the domain information, the model is able to propagate the label information from the source domain to the target domain. However, directly training on the multiple conditioned information is hard to converge. We propose to condition domain vectors and label vectors on different layers and train the model via alternating optimization. 

In this paper, we make the following contributions:
\begin{itemize}
\item We propose AlignGAN which is based on conditional GAN for aligning cross-domain images. We evaluate AlignGAN on numerous tasks and the experimental results demonstrate its effectiveness for aligning cross-domain images.
\item We also propose another model which conditions on multiple information such as domain information and label information. This model is able to propagate the label information from the source domain to the target domain. In addition, a 2-step alternating optimization algorithm is proposed to train this model.
\end{itemize}

\section{Related Works}
Goodfellow \etal \cite{Goodfellow2014} proposed the generative adversarial network (GAN) which has achieved great successes in generative models. After that, many works have been proposed to improve the image quality \cite{Mao2016, Radford2015, Zhang2016} or to stabilize the learning process \cite{Arjovsky2017, Metz2016, Salimans2016}. Further, GAN has been applied to various computer vision tasks such as image super-resolution \cite{Ledig2016}, text-to-image translation \cite{Reed2016}, and image-to-image translation \cite{Isola2016}. 

The most relevant work to this paper is CoGAN \cite{Liu2016} which also tries to align cross-domain images. In literature \cite{Liu2016}, the authors also tried to use conditional GAN for this task. However, their attempt failed in many tasks such as aligning digits and negative digits. Another task which is related to our work is image-to-image translation \cite{Kim2017, Zhu2017}. Both \cite{Zhu2016} and \cite{Kim2017} adopted two GANs which form a cycle mapping to form a reconstruction loss. Dong \etal \cite{Dong2017} proposed to use conditional GAN for image-to-image translation. They first trained a conditional GAN to learn shared features and then trained an encoder to map the images to latent vectors.

\section{Model}
In this section, we first briefly review GAN and conditional GAN in Section \ref{sec:gan}. Then we present the proposed AlignGAN in Section \ref{sec:aligngan}. Finally, the model to be conditioned on multiple information is introduced in Section \ref{sec:multiple_info}.

\subsection{GAN and Conditional GAN} \label{sec:gan}
The framework of GAN consists of two players, the discriminator $D$ and the generator $G$. Given a data distribution $p_\text{data}$, $G$ tries to learn the distribution $p_g$. $G$ starts from sampling noise input $\bm{z}$ from a uniform distribution $p_{z}(\bm{z})$, and then maps $\bm{z}$ to data space $G(\bm{z}; \theta_g)$. On the other hand, $D$ aims to distinguish whether a sample is from $p_\text{data}$ or from $p_g$.  The objective for GAN can be formulated as follows:

\begin{equation}
\label{eq:gan}
\begin{split}
\min_G \max_D V(D, G) = \mathbb{E}_{\bm{x} \sim p_{\text{data}}(\bm{x})}[\log D(\bm{x})]& \\
+ \mathbb{E}_{\bm{z} \sim p_{\bm{z}}(\bm{z})}[\log (1 - D(G(\bm{z})))]&.
\end{split}
\end{equation}

Conditional GAN introduces extra information $\bm{y}$ where both discriminator and generator are conditioned on $\bm{y}$. The objective for conditional GAN can be formulated as follows:

\begin{equation}
\label{eq:cgan}
\begin{split}
\min_G \max_D V(D, G) = \mathbb{E}_{\bm{x} \sim p_{\text{data}}(\bm{x})}[\log D(\bm{x}|\bm{y})]& \\
+ \mathbb{E}_{\bm{z} \sim p_{\bm{z}}(\bm{z})}[\log (1 - D(G(\bm{z}|\bm{y})))]&.
\end{split}
\end{equation}

\begin{figure}[t]
\centering
\begin{tabular}{cc}
 \includegraphics[width=1.5in]{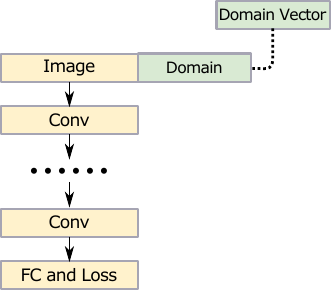}
&
 \includegraphics[width=1.5in]{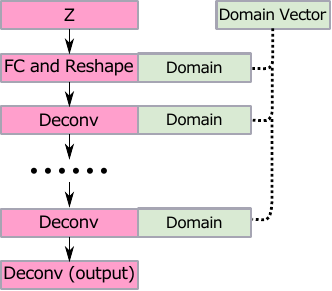}
\\
(a)
&
(b)
\end{tabular}
\caption{
Network architecture of AlignGAN. (a): The discriminator. (b): The generator. "Conv" and "Deconv" denote the convolutional layer and deconvolutional layer, respectively. "FC" denotes the fully connected layer. 
}
\label{fig:arch1}
\end{figure}

\subsection{AlignGAN} \label{sec:aligngan}

Our proposed AlignGAN is based on conditional GAN. The intuition is to learn the domain-specific semantics by the conditioned domain vectors and to learn the shared semantics by the other shared latent vectors. Previous attempt \cite{Liu2016} of using conditional GAN to align cross-domain images has shown its failure in many tasks. After extensive exploration, we conclude the following two rules for achieving successful learnings. 

First, for the generator, the noise input layer should not be conditioned by the domain vectors. Because the model should learn identical highest level semantics for different domains. For the other layers of the generator, they should be conditioned by the domain vectors. 

Second, for the discriminator, the image input layer should be conditioned by the domain vectors. Because the input layer generates the strongest signals to let the discriminator know which domain the images are from. For the other layers of the discriminator, we find that whether they are to be conditioned or not is not critical to the performance.

Based on the above two rules, we present the network architecture of AlignGAN in Figure \ref{fig:arch1}.

\begin{figure}[t]
\centering
\begin{tabular}{c}
 \includegraphics[width=2.18in]{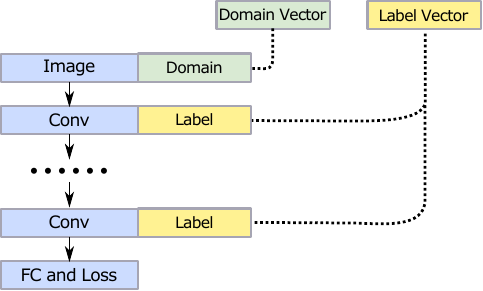}
 \\
(a)
\\
 \includegraphics[width=2.18in]{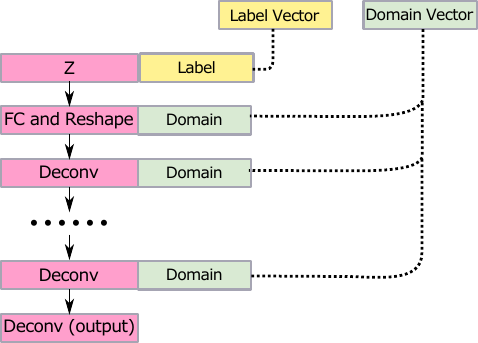}
\\
(b)
\end{tabular}
\caption{
Network architecture of the model conditioning on multiple information. (a): The discriminator. (b): The generator.
}
\label{fig:arch2}
\end{figure}

\subsection{Conditioning on Multiple Information} \label{sec:multiple_info}

Another model we proposed is to condition on multiple information such as domain information and label information. Domain information helps to align images from different domains and label information allows to control the class of generated images. One application of combining the two kinds of information is that we can propagate the label information from the source domain to the target domain when we only have the label information of source domain. The idea is to learn the semantics of label information from the source domain and to align the images from the domain information. As a result, the model is able to control the class of generated images of the target domain. One simple method is to concatenate the domain and label vectors first and then to be conditioned by the generator and discriminator. However, we find that this simple method is not able to converge. We propose to condition the domain vectors and label vectors separately, which means that the domain vectors and label vectors are conditioned by different layers. As stated in Section \ref{sec:aligngan}, the domain vectors should not be conditioned for the noise input layer of the generator. On the contrary, for the label vectors, the highest level semantics vary for different classes. Thus the label vectors should be conditioned by the noise input layer of the generator. As Figure \ref{fig:arch2} shows, we condition the label vectors on the layers which are not conditioned by the domain vectors.

\begin{figure*}[t]
\centering
\begin{tabular}{ccc}
 \includegraphics[width=0.307\textwidth]{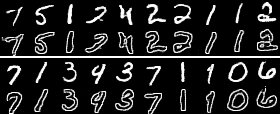}
&
 \includegraphics[width=0.307\textwidth]{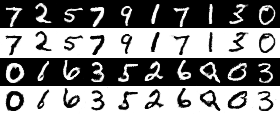}
& 
 \includegraphics[width=0.307\textwidth]{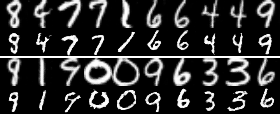}

\\
(a)
&
(b)
&
(c)
\end{tabular}
\caption{
Generated results on digit datasets. (a): Digits and edge digits. (b): Digits and negative digits. (c): USPS and MNIST.
}
\label{fig:digits}
\end{figure*}

\begin{figure*}[t]
\centering
\begin{tabular}{c}
 \includegraphics[width=0.97\textwidth]{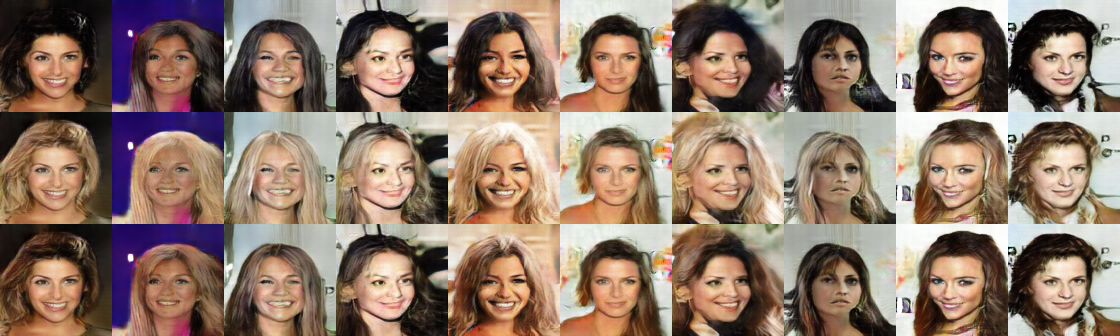}
 \\
 (a): Black hair, blond hair and brown hair.
 \\
 \includegraphics[width=0.97\textwidth]{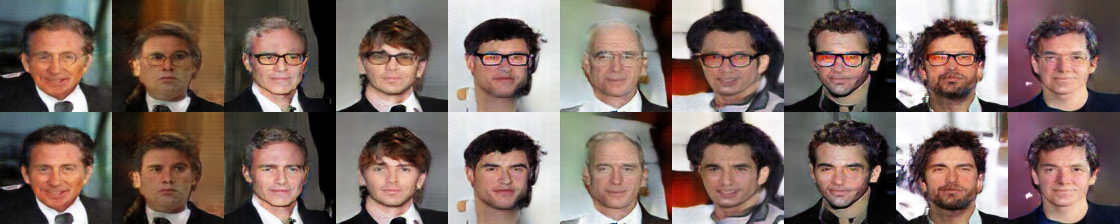}
 \\
  (b): With glasses and without glasses.
 \\
  \includegraphics[width=0.97\textwidth]{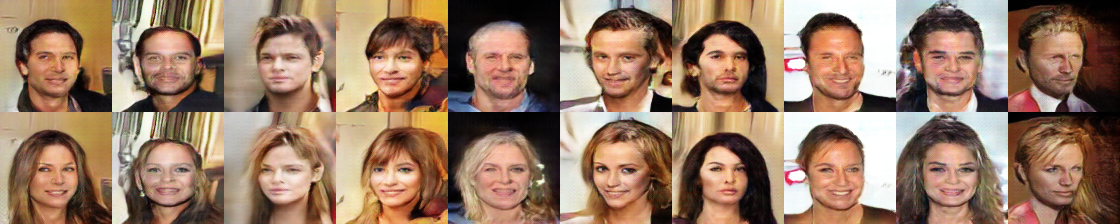}
 \\
  (c): Male and female.
 \\
  \includegraphics[width=0.97\textwidth]{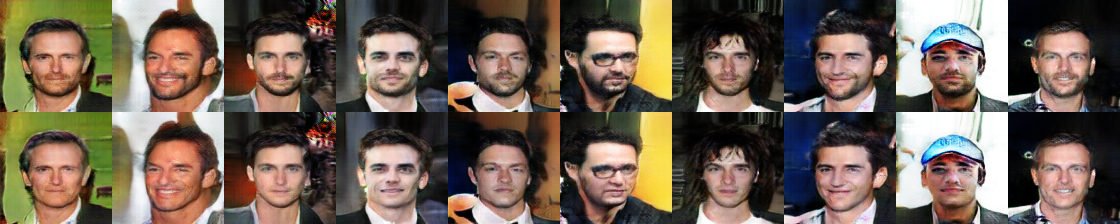}
 \\
  (d): With sideburns and without sideburns.
\end{tabular}
\caption{
Generated results on face dataset.
}
\label{fig:face}
\end{figure*}

\begin{figure*}[t]
\centering
\begin{tabular}{cc}
 \includegraphics[width=0.48\textwidth]{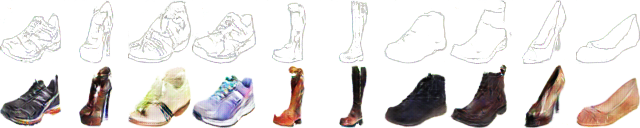}
&
 \includegraphics[width=0.48\textwidth]{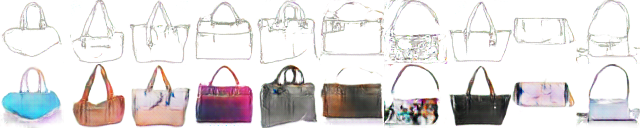}
\\
(a): Shoes.
&
(b): Handbags.
\end{tabular}
\caption{
Generated results on edge and photo dataset.
}
\label{fig:edge}
\end{figure*}

\vspace{2pt}
\noindent\textbf{2-Step Alternating Training.} We adopt a 2-step training algorithm to learn the domain-specific semantics and shared label semantics via alternating optimization. In the first step, we utilize the source domain images with label vectors to learn the label semantics, and the domain vectors are set to zero vectors. In the second step, we utilize both the source and target domain images with domain vectors to learn the domain-specific semantics, and the label vectors are set to zero vectors. The training procedure is formally presented in Algorithm \ref{alg:multiple_info}. Note that the hyperparameter $\tau$ is used to adjust the allocation of training iterations between domain semantics and label semantics. In our experiments, we set $\tau = 4$.

\begin{algorithm}[t]
\caption{\small Alternating training for conditioning on multiple information.
}
\begin{algorithmic}
\label{alg:multiple_info}

\small
\STATE \textbf{Input:} 
\STATE \INCIDENT $\bullet$ Source domain images $X_{s}$ with domain vectors $D_{s}$ and \\
\INCIDENT label vectors $L_{s}$.
\STATE \INCIDENT $\bullet$ Target domain images: $X_{t}$ with domain vectors $D_{t}$.

\FOR{number of training steps}
\IF{step mod $\tau == 0$}
    \STATE  $\bullet$ Update the discriminator using $X_{s}$ with $L_{s}$ and zero domain vectors.
    \STATE  $\bullet$ Update the generator with $L_{s}$ and zero domain vectors.
\ELSE
    \STATE  $\bullet$ Update the discriminator using $X_{s}$ and $X_{t}$ with $D_{s}$, $D_t$ and zero label vectors.
    \STATE  $\bullet$ Update the generator with $D_{s}$, $D_t$ and zero label vectors.
\ENDIF
\ENDFOR
\end{algorithmic}
\end{algorithm}

\section{Experiments}
\subsection{Implementation Details}
Except for the task of aligning digits and negative digits, we adopt LSGAN \cite{Mao2016} for training the models since LSGAN is able to generate higher quality images and stabilize the learning process. For the task of aligning digits and negative digits, we adopt regular GAN because we find that regular GAN performs well for this task while LSGAN will sometimes fail to align the images of digits and negative digits. For LSGAN, we select the parameters of $a=-1, b=1$, and $c=0$ which have been proven to minimize the Pearson $\chi^2$ divergence. Then Equation \ref{eq:gan} is replaced with the following formula:

\begin{equation}
\label{eq:lsgan}
\begin{split}
\min_D V_{\text{\tiny LSGAN}}(D) = &\frac{1}{2}\mathbb{E}_{\bm{x} \sim p_{\text{data}}(\bm{x})}\bigl[(D(\bm{x})-1)^2\bigr] \\
+ &\frac{1}{2}\mathbb{E}_{\bm{z} \sim p_{\bm{z}}(\bm{z})}\bigl[(D(G(\bm{z}))+1)^2\bigr] \\
\min_G V_{\text{\tiny LSGAN}}(G) = &\frac{1}{2}\mathbb{E}_{\bm{z} \sim p_{\bm{z}}(\bm{z})}\bigl[(D(G(\bm{z})))^2\bigr].
\end{split}
\end{equation}

We use Adam optimizer with learning rates of $0.0005$ for LSGAN and $0.0002$ for regular GAN. All the codes of our implementation will be public available soon.

\vspace{2pt}
\noindent\textbf{Model Selection} For LSGAN, we find that the quality of generated images will shift between good and bad during the training process. We select the model manually by checking the quality of generated images at some iterations.

\subsection{AlignGAN}
In this section, we evaluate AlignGAN on several datasets including digits, faces, edges, chairs, and cars. 
\subsubsection{Digits}
For this task, we use USPS and MNIST datasets to evaluate the performance of AlignGAN. Following literature \cite{Liu2016}, we first evaluate AlignGAN for the following two tasks. The first one is to align images of digits and edge digits. The second one is to align images of digits and negative digits. In addition, we further apply AlignGAN to align images of USPS and MNIST digits. As Figure \ref{fig:digits} shows, AlignGAN learns to align the images successfully for all the three tasks. 

\subsubsection{Faces}
We also evaluate AlignGAN on face images where the CelebFaces Attributes dataset \cite{Liu2015} is used for this experiment. We investigated the following four tasks: 1) alignment between different color hairs; 2) alignment between wearing eyeglasses and not wearing eyeglasses; 3) alignment between male and female; 4) alignment between males with sideburns and males without sideburns. The results are presented in Figure \ref{fig:face}, where the resolution of generated images is $112\times 112$.

\subsubsection{Edges and Photos}
Another evaluation is to align between edge images and realistic photos of handbags \cite{Zhu2016} or shoes \cite{Yu2014}. Figure \ref{fig:edge} shows the generated results with the resolution of $64 \times 64$ and we can observe that AlignGAN learns to align between edges and realistic photos successfully. 

\begin{figure}[h]
\centering
\begin{tabular}{c}
 \includegraphics[width=0.47 \textwidth]{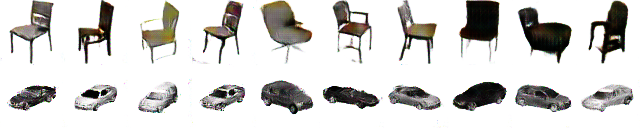}
\end{tabular}
\caption{
Generated results on chair and car dataset. The rotation angles of generated chairs and cars are highly correlated.
}
\label{fig:chair}
\end{figure}

\subsubsection{Chairs and Cars}
Following literature \cite{Kim2017}, we also investigate the task of aligning images of chairs \cite{Aubry2014} and cars \cite{Fidler2012} to study whether AlignGAN is able to learn the rotation relationship between the two different domains. As Figure \ref{fig:chair} shows, the rotation angles of generated chairs and cars are highly correlated.  

\begin{figure}[h]
\centering
\begin{tabular}{c}
 \includegraphics[width=0.46 \textwidth]{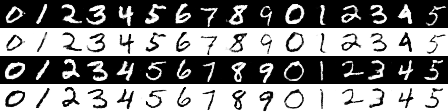}
 \\
 (a)

 \\
  \includegraphics[width=0.46 \textwidth]{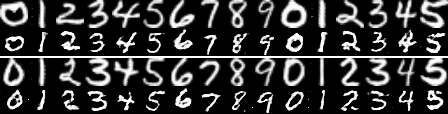}
 \\
 (b)

\end{tabular}
\caption{
Generated results on digit datasets conditioning on domain and label information. The digits are generated from $0$ to $9$ by controlling the label vectors. (a): Digits and negative digits. (b): USPS and MNIST.
}
\label{fig:multiple_negative}
\end{figure}

\subsection{Conditioning on Multiple Information}
We apply the proposed model conditioning on multiple information to two tasks. The MNIST dataset is used for the first task where the source and target domains are digits and negative digits, respectively. The second task is between USPS digits and MNIST digits. Only the label information of source domain is used during training. We generate the digits from $0$ to $9$ by controlling the label vectors and the results are shown in Figure \ref{fig:multiple_negative}. We have the following two observations. First, the paired images in Figure \ref{fig:multiple_negative} are highly correlated. Second, we are able to control the classes of generated target domain digits by adjusting the label vectors.

\section{Conclusions}
In this paper, we proposed two kinds of models. The first one called AlignGAN is for aligning cross-domain images based on conditional GAN. AlignGAN has been evaluated on numerous tasks and the experimental results demonstrate the effectiveness of AlignGAN for aligning cross-domain images. The second one is an extension of AlignGAN, which conditions on not only domain information but also label information. Conditioning on these two kinds of information, we are able to do label propagation from the source domain to the target domain.



{\small
\bibliographystyle{ieee}
\bibliography{aligngan}

\begin{thebibliography}{10}\itemsep=-1pt

\bibitem{Arjovsky2017}
M.~Arjovsky, S.~Chintala, and L.~Bottou.
\newblock Wasserstein gan.
\newblock {\em arXiv:1701.07875}, 2017.

\bibitem{Aubry2014}
M.~Aubry, D.~Maturana, A.~Efros, B.~Russell, and J.~Sivic.
\newblock Seeing 3d chairs: exemplar part-based 2d-3d alignment using a large
  dataset of cad models.
\newblock In {\em CVPR}, 2014.

\bibitem{Dong2017}
H.~Dong, P.~Neekhara, C.~Wu, and Y.~Guo.
\newblock Unsupervised image-to-image translation with generative adversarial
  networks.
\newblock {\em arXiv:1701.02676}, 2017.

\bibitem{Fidler2012}
S.~Fidler, S.~Dickinson, and R.~Urtasun.
\newblock 3d object detection and viewpoint estimation with a deformable 3d
  cuboid model.
\newblock In {\em Advances in Neural Information Processing Systems 25}, pages
  611--619. 2012.

\bibitem{Goodfellow2014}
I.~Goodfellow, J.~Pouget-Abadie, M.~Mirza, B.~Xu, D.~Warde-Farley, S.~Ozair,
  A.~Courville, and Y.~Bengio.
\newblock Generative adversarial nets.
\newblock In {\em Advances in Neural Information Processing Systems (NIPS)},
  pages 2672--2680, 2014.

\bibitem{Isola2016}
P.~Isola, J.-Y. Zhu, T.~Zhou, and A.~A. Efros.
\newblock Image-to-image translation with conditional adversarial networks.
\newblock {\em arXiv:1611.07004}, 2016.

\bibitem{Kim2017}
T.~Kim, M.~Cha, H.~Kim, J.~K. Lee, and J.~Kim.
\newblock Learning to discover cross-domain relations with generative
  adversarial networks.
\newblock {\em arXiv:1703.05192}, 2017.

\bibitem{Ledig2016}
C.~Ledig, L.~Theis, F.~Huszar, J.~Caballero, A.~Cunningham, A.~Acosta,
  A.~Aitken, A.~Tejani, J.~Totz, Z.~Wang, and W.~Shi.
\newblock {Photo-Realistic Single Image Super-Resolution Using a Generative
  Adversarial Network}.
\newblock {\em arXiv:1609.04802}, 2016.

\bibitem{Liu2016}
M.-Y. Liu and O.~Tuzel.
\newblock Coupled generative adversarial networks.
\newblock In {\em Advances in Neural Information Processing Systems (NIPS)},
  pages 469--477, 2016.

\bibitem{Liu2015}
Z.~Liu, P.~Luo, X.~Wang, and X.~Tang.
\newblock Deep learning face attributes in the wild.
\newblock In {\em Proceedings of International Conference on Computer Vision
  (ICCV)}, 2015.

\bibitem{Mao2016}
X.~Mao, Q.~Li, H.~Xie, R.~Y. Lau, Z.~Wang, and S.~P. Smolley.
\newblock Least squares generative adversarial networks.
\newblock {\em arXiv:1611.04076}, 2016.

\bibitem{Metz2016}
L.~Metz, B.~Poole, D.~Pfau, and J.~Sohl-Dickstein.
\newblock Unrolled generative adversarial networks.
\newblock {\em arXiv:1611.02163}, 2016.

\bibitem{Mirza2014}
M.~Mirza and S.~Osindero.
\newblock {Conditional Generative Adversarial Nets}.
\newblock {\em arXiv:1411.1784}, 2014.

\bibitem{Radford2015}
A.~Radford, L.~Metz, and S.~Chintala.
\newblock Unsupervised representation learning with deep convolutional
  generative adversarial networks.
\newblock In {\em International Conference on Learning Representations (ICLR)},
  2015.

\bibitem{Reed2016}
S.~Reed, Z.~Akata, X.~Yan, L.~Logeswaran, B.~Schiele, and H.~Lee.
\newblock Generative adversarial text-to-image synthesis.
\newblock In {\em Proceedings of The 33rd International Conference on Machine
  Learning (ICML)}, 2016.

\bibitem{Salimans2016}
T.~Salimans, I.~Goodfellow, W.~Zaremba, V.~Cheung, A.~Radford, X.~Chen, and
  X.~Chen.
\newblock Improved techniques for training gans.
\newblock In {\em Advances in Neural Information Processing Systems (NIPS)},
  pages 2226--2234, 2016.

\bibitem{Wang2012}
S.~Wang, L.~Zhang, Y.~Liang, and Q.~Pan.
\newblock Semi-coupled dictionary learning with applications to image
  super-resolution and photo-sketch synthesis.
\newblock In {\em Computer Vision and Pattern Recognition (CVPR)}, pages
  2216--2223, June 2012.

\bibitem{Yu2014}
A.~Yu and K.~Grauman.
\newblock {F}ine-{G}rained {V}isual {C}omparisons with {L}ocal {L}earning.
\newblock In {\em Computer Vision and Pattern Recognition (CVPR)}, June 2014.

\bibitem{Zhang2016}
H.~Zhang, T.~Xu, H.~Li, S.~Zhang, X.~Huang, X.~Wang, and D.~Metaxas.
\newblock Stackgan: Text to photo-realistic image synthesis with stacked
  generative adversarial networks.
\newblock {\em arXiv:1612.03242}, 2016.

\bibitem{Zhu2016}
J.-Y. Zhu, P.~Kr{\"a}henb{\"u}hl, E.~Shechtman, and A.~A. Efros.
\newblock Generative visual manipulation on the natural image manifold.
\newblock In {\em Proceedings of European Conference on Computer Vision
  (ECCV)}, 2016.

\bibitem{Zhu2017}
J.-Y. Zhu, T.~Park, P.~Isola, and A.~A. Efros.
\newblock Unpaired image-to-image translation using cycle-consistent
  adversarial networks.
\newblock {\em arXiv:1703.10593}, 2017.

\end{thebibliography}
}

\end{document}